\pdfoutput=1

\def\year{2021}\relax
\documentclass[letterpaper]{article} 
\usepackage{aaai21}  
\usepackage{times}  
\usepackage{helvet} 
\usepackage{courier}  
\usepackage[hyphens]{url}  
\usepackage{graphicx} 
\usepackage{natbib}  
\usepackage{caption} 
\usepackage{amssymb,amsmath}
\usepackage{algorithm}
\usepackage{algorithmic}

\def \f {\tilde{f}}
\def \R {\mathbb{R}}
\def \K {\mathcal{K}}
\def \E {\mathbb{E}}
\def \x {\mathbf{x}}
\def \y {\mathbf{y}}
\def \z {\mathbf{z}}
\def \v {\mathbf{v}}

\DeclareMathOperator*{\argmin}{argmin}
\DeclareMathOperator*{\ii}{in}
\DeclareMathOperator*{\oo}{out}
\newtheorem{thm}{Theorem}
\newtheorem{lem}{Lemma}
\newtheorem{myDef}{Definition}
\newtheorem{assum}{Assumption}

\title{Projection-free Online Learning over Strongly Convex Sets}
\author{
    Yuanyu Wan\textsuperscript{\rm 1},~ Lijun Zhang\textsuperscript{\rm 1,\rm 2,}\thanks{Lijun Zhang is the corresponding author.}
}
\affiliations{

    \textsuperscript{\rm 1}National Key Laboratory for Novel Software Technology, Nanjing University, Nanjing 210023, China\\
    \textsuperscript{\rm 2}Pazhou Lab, Guangzhou 510330, China\\
    \{wanyy, zhanglj\}@lamda.nju.edu.cn
}

\begin{document}

\maketitle

\begin{abstract}
To efficiently solve online problems with complicated constraints, projection-free algorithms including online frank-wolfe (OFW) and its variants have received significant interest recently. However, in the general case, existing efficient projection-free algorithms only achieved the regret bound of $O(T^{3/4})$, which is worse than the regret of projection-based algorithms, where $T$ is the number of decision rounds. In this paper, we study the special case of online learning over strongly convex sets, for which we first prove that OFW can enjoy a better regret bound of $O(T^{2/3})$ for general convex losses. The key idea is to refine the decaying step-size in the original OFW by a simple line search rule. Furthermore, for strongly convex losses, we propose a strongly convex variant of OFW by redefining the surrogate loss function in OFW. We show that it achieves a regret bound of $O(T^{2/3})$ over general convex sets and a better regret bound of $O(\sqrt{T})$ over strongly convex sets.
\end{abstract}

\section{Introduction}
Online convex optimization (OCO) is a powerful framework that has been used to model and solve problems from diverse domains such as online routing \citep{Awerbuch04,Awerbuch2008} and online portfolio selection \citep{Blum1999,Portfolio_Agarwal,Portfolio_Luo}. 
In OCO, an online player plays a repeated game of $T$ rounds against an adversary. At each round $t$, the player chooses a decision $\x_t$ from a convex set $\K$. After that, a convex function $f_t(\x):\K\to\R$ chosen by the adversary is revealed, and incurs a loss $f_t(\x_t)$ to the player. The goal of the player is to choose decisions so that the regret defined as
\[R(T)=\sum_{t=1}^Tf_t(\x_t)-\min_{\x\in\K}\sum_{t=1}^Tf_t(\x)\]
is minimized. Over the past decades, various algorithms such as online gradient descent (OGD) \citep{Zinkevich2003}, online Newton step \citep{Hazan_2007} and follow-the-regularized-leader \citep{ShaiThesis,Shai07} have been proposed to yield optimal regret bounds under different scenarios. 

To ensure the feasibility of each decision, a common way in these algorithms is to apply a projection operation to any infeasible decision. However, in many high-dimensional problems with complicated decision sets, the projection step becomes a computational bottleneck \citep{Zhang-ICML13,chen-UAI2016,Yang-ICML17}. 
For example, in semi-definite programs \citep{FW_Hazan08} and multiclass classification \citep{SVFW}, it amounts to computing the singular value decomposition (SVD) of a matrix, when the decision set consists of all matrices with bounded trace norm. To tackle the computational bottleneck, \citet{Hazan2012} proposed a projection-free algorithm called online Frank-Wolfe (OFW) that replaces the time-consuming projection step with a more efficient linear optimization step. Taking matrix completion as an example again, the linear optimization step can be carried out by computing the top singular vector pair of a matrix, which is significantly faster than computing the SVD. Although OFW can efficiently handle complicated decision sets, it only attains a regret bound of $O(T^{3/4})$ for the general OCO, which is worse than the optimal $O(\sqrt{T})$ regret bound achieved by projection-based algorithms.
\begin{table*}
  \centering
  \begin{tabular}{|c|c|c|c|}
     \hline
     Algorithm & Extra Condition on Loss & Extra Condition on $\K$  & Regret Bound\\
     \hline
     OFW & & & $O(T^{3/4})$\\
     \hline
     LLOO-OCO & &polyhedral & $O(\sqrt{T})$\\
     \hline
     LLOO-OCO & strongly convex &polyhedral & $O(\log T)$\\
     \hline
     Fast OGD & & smooth & $O(\sqrt{T})$\\
     \hline
     Fast OGD &strongly convex & smooth& $O(\log T)$\\
     \hline
     OSPF &smooth  & & $O(T^{2/3})$\\
     \hline
     OFW with Line Search (this work) & & strongly convex & $O(T^{2/3})$ \\
     \hline
     SC-OFW (this work) & strongly convex &  & $O(T^{2/3})$ \\
     \hline
     SC-OFW (this work) & strongly convex& strongly convex & $O(\sqrt{T})$ \\
     \hline
  \end{tabular}
  \caption{Comparisons of regret bounds for efficient projection-free online algorithms including OFW \citep{Hazan2012,Hazan2016}, LLOO-OCO \citep{Garber16}, Fast OGD \citep{kevy_smooth}, OSPF \citep{Hazan20} and our algorithms.} \label{table1}
\end{table*}

More specifically, OFW is an online extension of an offline algorithm called Frank-Wolfe (FW) \citep{FW-56,Revist_FW} that iteratively performs linear optimization steps to minimize a convex and smooth function. In each round, OFW updates the decision by utilizing a single step of FW to minimize a surrogate loss function, which implies that the approximation error caused by the single step of FW could be the main reason for the regret gap between projection-based algorithms and OFW. Recently, \citet{Gaber_ICML_15} made a quadratic improvement in the convergence rate of FW for strongly convex and smooth offline optimization over strongly convex sets compared to the general case. They used a simple line search rule to choose the step-size of FW, which allows FW to converge faster even if the strong convexity of the decision set is unknown. It is therefore natural to ask whether the faster convergence of FW can be utilized to improve the regret of OFW. In this paper, we give an affirmative answer by improving OFW to achieve an regret bound of $O(T^{2/3})$ over strongly convex sets, which is better than the original $O(T^{3/4})$ regret bound. Inspired by \citet{Gaber_ICML_15}, the key idea is to refine the decaying step-size in the original OFW by a simple line search rule.

Furthermore, we study OCO with strongly convex losses, and propose a strongly convex variant of OFW (SC-OFW). Different from OFW, we introduce a new surrogate loss function from \citet{Garber16} to utilize the strong convexity of losses, which has been used to achieve an $O(\log T)$ regret bound for strongly convex OCO over polyhedral sets. Theoretical analysis reveals that SC-OFW for strongly convex OCO attains a regret bound of $O(T^{2/3})$ over general convex sets\footnote{When this paper is under review, we notice that a concurrent work \citep{Garber_SOFW} also proposed an algorithm similar to SC-OFW and established the $O(T^{2/3})$ regret bound.} and a better regret bound of $O(\sqrt{T})$ over strongly convex sets.
\section{Related Work}
In this section, we briefly review the related work about efficient projection-free algorithms for OCO, the time complexity of which is a constant in each round.

OFW \citep{Hazan2012,Hazan2016} is the first projection-free algorithms for OCO, which attains a regret bound of $O(T^{3/4})$ by only performing $1$ linear optimization step at each round. Recently, some studies have proposed projection-free online algorithms which are as efficient as OFW and attain better regret bounds, for special cases of OCO. If the decision set is a polytope, \citet{Garber16} proposed a variant of OFW named as LLOO-based online convex optimization (LLOO-OCO), which enjoy an $O(\sqrt{T})$ regret bound for convex losses and an $O(\log T)$ regret bound for strongly convex losses. For OCO over smooth sets, \citet{kevy_smooth} proposed a projection-free variant of OGD via devising a fast approximate projection for such sets, and established $O(\sqrt{T})$ and $O(\log T)$ regret bounds for convex and strongly convex losses, respectively. Besides these improvements for OCO over special decision sets, \citet{Hazan20} proposed online smooth projection free algorithm (OSPF) for OCO with smooth losses, which is a randomized method and achieves an expected regret bound of $O(T^{2/3})$.

Furthermore, OFW has been extended to two practical scenarios. The first scenario is the bandit setting \citep{OBO05,Bubeck15}, where only the loss value is available to the player. \citet{chen19} proposed the first bandit variant of OFW, and established an expected regret bound of $O(T^{4/5})$. Later, two improved bandit variants of OFW were proposed to enjoy better expected regret bounds on the order of $O(T^{3/4})$ for convex losses \citep{Garber19} and $\widetilde{O}(T^{2/3})$ for strongly convex losses \citep{Garber_SOFW}. The second scenario is the distributed setting \citep{DADO2011,D-ODA}, where many players are distributed over a network and each player can only access to the local loss function. The first projection-free algorithm for distributed OCO was proposed by \citet{wenpeng17}, which requires the communication complexity of $O(T)$. Recently, \citet{Wan-ICML-2020} further reduced the communication complexity from $O(T)$ to $O(\sqrt{T})$.

Despite these great progresses about projection-free online learning, for the general OCO, OFW is still the best known efficient projection-free algorithm and the regret bound of $O(T^{3/4})$ has remained unchanged. In this paper, we will study OCO over strongly convex sets, and improve the regret bound to $O(T^{2/3})$ and $O(\sqrt{T})$ for convex and strongly convex losses, respectively. The detailed comparisons between efficient projection-free algorithms are summarized in Table \ref{table1}.

\section{Main Results}
In this section, we first introduce necessary preliminaries including common notations, definitions and assumptions. Then, we present an improved regret bound for OFW over strongly convex sets by utilizing the line search. Finally, we introduce our SC-OFW algorithm for strongly convex OCO as well as its theoretical guarantees.
\subsection{Preliminaries}
In this paper, the convex set $\K$ belongs to a finite vector space $\E$. 
For any $\x,\y\in\K$, the inner product is denoted by $\langle\x,\y\rangle$. We recall two standard definitions for smooth and strongly convex functions \citep{Boyd04}, respectively.
\begin{myDef}
\label{def1}
Let $f(\x):\K\to\mathbb{R}$ be a function over $\K$. It is called $\beta$-smooth over $\K$ if for all $\x,\mathbf{y}\in \K$
\[f(\mathbf{y})\leq f(\x)+\langle\nabla f(\x),\mathbf{y}-\x\rangle+\frac{\beta}{2}\|\mathbf{y}-\x\|_2^2.\]
\end{myDef}
\begin{myDef}
\label{def2}
Let $f(\x):\K\to\mathbb{R}$ be a function over $\K$. It is called $\alpha$-strongly convex over $\K$ if for all $\x,\mathbf{y}\in \K$
\[f(\mathbf{y})\geq f(\x)+\langle\nabla f(\x),\mathbf{y}-\x\rangle+\frac{\alpha}{2}\|\mathbf{y}-\x\|_2^2.\]
\end{myDef}
When $f(\x):\K\to\mathbb{R}$ is an $\alpha$-strongly convex function and $\x^\ast=\argmin_{\x\in\K} f(\x)$, \citet{Gaber_ICML_15} have proved that for any $\x\in\K$
\begin{equation}
\label{cor_scvx}
\frac{\alpha}{2}\|\x-\x^\ast\|_2^2\leq f(\x)-f(\x^\ast)
\end{equation}
and
 \begin{equation}
\label{dual_cor_scvx}
\|\nabla f(\x)\|_2\geq\sqrt{\frac{\alpha}{2}}\sqrt{f(\x)-f(\x_\ast)}.
\end{equation}
Then, we introduce the definition of the strongly convex set, which has been well studied in offline optimization \citep{Levitin66,Demyanov70,Dunn79,Gaber_ICML_15,Jarrid19}.
\begin{myDef}
\label{def3}
A convex set $\K\in\E$ is called $\alpha$-strongly convex with respect to a norm $\|\cdot\|$ if for any $\x,\y\in\K$, $\gamma\in[0,1]$ and $\z\in\E$ such that $\|\z\|=1$, it holds that
\[
\gamma\x+(1-\gamma)\y+\gamma(1-\gamma)\frac{\alpha}{2}\|\x-\y\|^2\z \in\K.
\]
\end{myDef}
As shown by \citet{Gaber_ICML_15}, various balls induced by $\ell_p$ norms, Schatten norms and group
norms are strongly convex, which are commonly used to constrain the decision. For example, the $\ell_p$ norm ball defined as \[\K=\{\x\in\mathbb{R}^d|\|\x\|_p\leq r\}\] is $\frac{(p-1)d^{\frac{1}{2}-\frac{1}{p}}}{r}$-strongly convex with respect to the $\ell_2$ norm for any $p\in(1,2]$ (\citeauthor{Gaber_ICML_15}, \citeyear{Gaber_ICML_15}, Corollary 1).

We also introduce two common assumptions in OCO \citep{Online:suvery,Hazan2016}.
\begin{assum}
\label{assum1}
The diameter of the convex decision set $\K$ is bounded by $D$, i.e., \[\|\x-\mathbf{y}\|_2\leq D\] for any $\x,\mathbf{y}\in \K$.
\end{assum}
\begin{assum}
\label{assum2}
At each round $t$, the loss function $f_t(\x)$ is $G$-Lipschitz over $\K$, i.e., \[|f_t(\x)-f_t(\mathbf{y})|\leq G\|\x-\mathbf{y}\|_2\]  for any $\x,\mathbf{y}\in \K$.
\end{assum}
\begin{algorithm}[t]
\caption{OFW with Line Search}
\label{OFW}
\begin{algorithmic}[1]
\STATE \textbf{Input:} feasible set $\mathcal{K}$, $\eta$
\STATE \textbf{Initialization:} choose $\x_{1}\in\K$
\FOR{$t=1,\cdots,T$}
\STATE Define $F_t(\x)=\eta\sum_{\tau=1}^t\langle\nabla f_{\tau}(\x_\tau),\x\rangle+\|\x-\x_1\|_2^2$
\STATE $\mathbf{v}_{t}\in\argmin\limits_{\mathbf{x}\in\mathcal{K}} \langle\nabla F_t(\x_{t}),\mathbf{x}\rangle$
\STATE $\sigma_t=\argmin\limits_{\sigma\in[0,1]}\langle\sigma(\v_t-\x_t),\nabla F_t(\x_t)\rangle+\sigma^2\|\v_t-\x_t\|_2^2$
\STATE $\x_{t+1}=\x_{t}+\sigma_t(\mathbf{v}_{t}-\mathbf{x}_{t})$
\ENDFOR
\end{algorithmic}
\end{algorithm}
\subsection{OFW with Line Search}
For the general OCO, OFW \citep{Hazan2012,Hazan2016} arbitrarily chooses $\x_1$ from $\K$, and then iteratively updates its decision as the following linear optimization step
\begin{equation*}
\begin{split}
&\v=\argmin_{\x\in\K}\langle\nabla F_{t}(\x_t),\x\rangle\\
&\x_{t+1}=\x_t+\sigma_t(\v-\x_t)
\end{split}
\end{equation*}
where \begin{equation}
\label{eq_F1}
F_t(\x)=\eta\sum_{\tau=1}^t\langle\nabla f_{\tau}(\x_\tau),\x\rangle+\|\x-\x_1\|_2^2
\end{equation} is the surrogate loss function, $\sigma_t$ and $\eta$ are two parameters. According to \citet{Hazan2016}, OFW with $\eta=O(T^{-3/4})$ and $\sigma_t=O(t^{-1/2})$ attains the regret bound of $O(T^{3/4})$. However, this decaying step-size $\sigma_t=O(t^{-1/2})$ cannot utilize the property of strongly convex sets. Inspired by a recent progress on FW over strongly convex sets \citep{Gaber_ICML_15}, we utilized a line search rule to choose the parameter $\sigma_t$ as
\[\sigma_t=\argmin_{\sigma\in[0,1]}\langle\sigma(\v_t-\x_t),\nabla F_t(\x_t)\rangle+\sigma^2\|\v_t-\x_t\|_2^2.\]
The detailed procedures are summarized in Algorithm~\ref{OFW}.

From previous discussions, the approximation error of minimizing $F_t(\x)$ by a single step of FW has a significant impact on the regret of OFW. Therefore, we first present the following lemma with respect to the approximation error for Algorithm~\ref{OFW} over strongly convex sets.
\begin{lem}
\label{thm1_lem1}
Let $\K$ be an $\alpha_K$-strongly convex set with respect to the $\ell_2$ norm. Let $\x_{t}^\ast=\argmin_{\x\in \K}F_{t-1}(\x)$ for any $t\in[T+1]$, where $F_t(\x)$ is defined in (\ref{eq_F1}). Then, for any $t\in[T+1]$, Algorithm~\ref{OFW} with $\eta=\frac{D}{2G(T+2)^{2/3}}$ has
\[F_{t-1}(\x_{t})-F_{t-1}(\x_t^\ast)\leq\epsilon_t=\frac{C}{(t+2)^{2/3}}\]
where $C=\max\left(4D^2,\frac{4096}{3\alpha_K^2}\right).$
\end{lem}
We find that the approximation error incurred by a single step of FW is upper bounded by $O(t^{-2/3})$ for Algorithm~\ref{OFW} over strongly convex sets. For a clear comparison, we note that the approximation error for the original OFW over general convex sets has a worse bound of $O(1/\sqrt{t})$ (\citeauthor{Hazan2016}, \citeyear{Hazan2016}, Lemma 7.3).

By applying Lemma~\ref{thm1_lem1} and further analyzing the property of decisions $\x_{1}^\ast,\cdots,\x_{T}^\ast$ defined in Lemma~\ref{thm1_lem1}, we prove that OFW with line search enjoys the following regret bound over strongly convex sets.
\begin{thm}
\label{thm1}
Let $\K$ be an $\alpha_K$-strongly convex set with respect to the $\ell_2$ norm and $C=\max\left(4D^2,\frac{4096}{3\alpha_K^2}\right)$. For any $\x^\ast\in\K$, Algorithm~\ref{OFW} with $\eta=\frac{D}{2G(T+2)^{2/3}}$ ensures
\[
\sum_{t=1}^Tf_t(\x_t)-\sum_{t=1}^Tf_t(\x^\ast)\leq\frac{11}{4}G\sqrt{C}(T+2)^{2/3}.
\]
\end{thm}
We find that the $O(T^{2/3})$ regret bound in Theorem~\ref{thm1} is better than the $O(T^{3/4})$ bound achieved by previous studies for the original OFW over general convex sets \citep{Hazan2012,Hazan2016}.
\begin{algorithm}[t]
\caption{Strongly Convex Variant of OFW (SC-OFW)}
\label{SC_OFW}
\begin{algorithmic}[1]
\STATE \textbf{Input:} feasible set $\mathcal{K}$, $\lambda$
\STATE \textbf{Initialization:} choose $\x_{1}\in\K$
\FOR{$t=1,\cdots,T$}
\STATE Define $F_t(\x)=\sum_{\tau=1}^t(\langle\nabla f_{\tau}(\x_\tau),\x\rangle+\frac{\lambda}{2}\|\x-\x_\tau\|_2^2)$
\STATE $\mathbf{v}_{t}\in\argmin\limits_{\mathbf{x}\in\mathcal{K}} \langle\nabla F_t(\x_{t}),\mathbf{x}\rangle$
\STATE $\sigma_t=\argmin\limits_{\sigma\in[0,1]}\langle\sigma(\v_t-\x_t),\nabla F_t(\x_t)\rangle+\frac{\sigma^2\lambda t}{2}\|\v_t-\x_t\|_2^2$
\STATE $\x_{t+1}=\x_{t}+\sigma_t(\mathbf{v}_{t}-\mathbf{x}_{t})$
\ENDFOR
\end{algorithmic}
\end{algorithm}
\subsection{SC-OFW for Strongly Convex Losses}
Moreover, we propose a strongly convex variant of OFW (SC-OFW), which is inspired by \citet{Garber16}. To be precise, \citet{Garber16} have proposed a variant of OFW for strongly convex OCO over polyhedral sets, which enjoys the $O(\log T)$ regret bound. Compared with the original OFW, their algorithm has two main differences. First, a local linear optimization step for polyhedral sets is developed to replace the linear optimization step utilized in OFW. Second, to handle $\lambda$-strongly convex losses, $F_t(\x)$ is redefined as \begin{equation}
\label{F_poly}
\begin{split}
F_t(\x)=&\sum_{\tau=1}^t\left(\langle\nabla f_{\tau}(\x_\tau),\x\rangle+\frac{\lambda}{2}\|\x-\x_\tau\|_2^2\right)\\
&+\frac{c\lambda}{2}\|\x-\x_1\|_2^2
\end{split}
\end{equation}
where $c$ is a parameter that depends on properties of polyhedral sets.

Since this paper considers OCO over strongly convex sets, our SC-OFW adopts the linear optimization step utilized in the original OFW, and simplifies $F_t(\x)$ in (\ref{F_poly}) as \begin{equation}
\label{def_F2}
F_t(\x)=\sum_{\tau=1}^t\left(\langle\nabla f_{\tau}(\x_\tau),\x\rangle+\frac{\lambda}{2}\|\x-\x_\tau\|_2^2\right)
\end{equation}
by simply setting $c=0$. The detailed procedures are summarized in Algorithm~\ref{SC_OFW}, where the parameter $\sigma_t$ is selected by the line search as
\[\sigma_t=\argmin_{\sigma\in[0,1]}\langle\sigma(\v_t-\x_t),\nabla F_t(\x_t)\rangle+\frac{\sigma^2\lambda t}{2}\|\v_t-\x_t\|_2^2.\]
To the best of our knowledge, SC-OFW is the first projection-free algorithm for strongly convex OCO over any decision set. 

In the following lemma, we upper bound the approximation error of minimizing the surrogate loss function for SC-OFW over strongly convex sets.
\begin{lem}
\label{thm2_lem1}
Let $\K$ be an $\alpha_K$-strongly convex set with respect to the $\ell_2$ norm. Let $\x_{t}^\ast=\argmin_{\x\in \K}F_{t-1}(\x)$ for any $t=2,\cdots,T+1$, where $F_t(\x)$ is defined in (\ref{def_F2}). Then, for any $t=2,\cdots,T+1$, Algorithm~\ref{SC_OFW} has
\[F_{t-1}(\x_{t})-F_{t-1}(\x_t^\ast)\leq C\]
where $C=\max\left(\frac{4(G+\lambda D)^2}{\lambda},\frac{288\lambda}{\alpha_K^2}\right).$
\end{lem}
Furthermore, based on Lemma~\ref{thm2_lem1}, we provide the following theorem with respect to the regret bound of SC-OFW over strongly convex sets.
\begin{thm}
\label{thm2}
Let $\K$ be an $\alpha_K$-strongly convex set with respect to the $\ell_2$ norm. If $f_t(\x)$ is $\lambda$-strongly convex for any $t\in[T]$, for any $\x^\ast\in\K$, Algorithm~\ref{SC_OFW} ensures
\[
\sum_{t=1}^Tf_t(\x_t)-\sum_{t=1}^Tf_t(\x^\ast)\leq C\sqrt{2T}+\frac{C\ln T}{2}+GD
\]
where $C=\max\left(\frac{4(G+\lambda D)^2}{\lambda},\frac{288\lambda}{\alpha_K^2}\right).$
\end{thm}
From Theorem~\ref{thm2}, we achieve a regret bound of $O(\sqrt{T})$ for strongly convex losses, which is better than the above $O(T^{2/3})$ bound for general convex losses.

Moreover, we show the regret bound of Algorithm~\ref{SC_OFW} over any general convex set $\K$ in the following theorem.
\begin{thm}
\label{thm3}
If $f_t(\x)$ is $\lambda$-strongly convex for any $t\in[T]$, for any $\x^\ast\in\K$, Algorithm~\ref{SC_OFW} ensures
\[
\sum_{t=1}^Tf_t(\x_t)-\sum_{t=1}^Tf_t(\x^\ast)\leq \frac{3\sqrt{2}CT^{2/3}}{8}+\frac{C\ln T}{8}+GD
\]
where $C=\frac{16(G+\lambda D)^2}{\lambda}.$
\end{thm}
By comparing Theorem~\ref{thm3} with Theorem~\ref{thm2}, we can find that our Algorithm~\ref{SC_OFW} enjoys a better regret bound when the decision set is strongly convex.
\section{Theoretical Analysis}
In this section, we prove Theorems \ref{thm1} and \ref{thm2}. The omitted proofs can be found in the appendix.
\subsection{Proof of Theorem~\ref{thm1}}
In the beginning, we define $\x_{t}^\ast=\argmin_{\x\in \K}F_{t-1}(\x)$ for any $t\in[T+1]$, where $F_t(\x)$ is defined in (\ref{eq_F1}). 

Since each function $f_t(\x)$ is convex, we have
\begin{equation}
\label{thm1_eq1}
\begin{split}
&\sum_{t=1}^Tf_t(\x_t)-\sum_{t=1}^Tf_t(\x^\ast)\\
\leq&\sum_{t=1}^T\langle\nabla f_t(\x_t),\x_t-\x^\ast\rangle\\
=&\underbrace{\sum_{t=1}^T\langle\nabla f_t(\x_t),\x_t-\x_t^\ast\rangle}_{:=A}+\underbrace{\sum_{t=1}^T\langle\nabla f_t(\x_t),\x_t^\ast-\x^\ast\rangle}_{:=B}.\\
\end{split}
\end{equation}
So, we will upper bound the regret by analyzing $A$ and $B$.

By applying Lemma~\ref{thm1_lem1}, we have
\begin{equation}
\label{thm1_eq5}
\begin{split}
A=&\sum_{t=1}^T\langle\nabla f_t(\x_t),\x_t-\x_t^\ast\rangle\\
\leq&\sum_{t=1}^T\|\nabla f_t(\x_t)\|_2\|\x_t-\x_t^\ast\|_2\\
\leq&\sum_{t=1}^TG\sqrt{F_{t-1}(\x_t)-F_{t-1}(\x_t^\ast)}\\
\leq&\sum_{t=1}^T\frac{G\sqrt{C}}{(t+2)^{1/3}}\\
\leq&\frac{3G\sqrt{C}(T+2)^{2/3}}{2}
\end{split}
\end{equation}
where the second inequality is due to (\ref{cor_scvx}) and the fact that $F_{t-1}(\x)$ is $2$-strongly convex, and the last inequality is due to $\sum_{t=1}^T{(t+2)^{-1/3}}\leq3(T+2)^{2/3}/2$.

To bound $B$, we introduce the following lemma.
\begin{lem}
\label{lem_garber16}
(Lemma 6.6 of \citet{Garber16})
Let $\{f_t(\x)\}_{t=1}^T$ be a sequence of loss functions and let $\x_t^\ast\in\argmin_{\x\in\K}\sum_{\tau=1}^tf_{\tau}(\x)$ for any $t\in[T]$. Then, it holds that
\[\sum_{t=1}^Tf_t(\x_t^\ast)-\min_{\x\in\K}\sum_{t=1}^Tf_t(\x)\leq0.\]
\end{lem}
To apply Lemma~\ref{lem_garber16}, we define $\tilde{f}_1(\x)=\langle\eta\nabla f_1(\x_1),\x\rangle+\|\x-\x_1\|_2^2$ for $t=1$ and $\tilde{f}_t(\x)=\langle\eta\nabla f_t(\x_{t}),\x\rangle$ for any $t\geq2$. We recall that $F_{t}(\x)=\sum_{\tau=1}^t\f_\tau(\x)$ and $\x_{t+1}^\ast=\argmin_{\x\in \K}F_{t}(\x)$ for any $t=1,\cdots,T$. Then, by applying Lemma~\ref{lem_garber16} to $\{\f_t(\x)\}_{t=1}^T$, we have
\begin{equation*}
\begin{split}
\sum_{t=1}^T\f_t(\x_{t+1}^\ast)-\sum_{t=1}^T\f_t(\x^\ast)\leq 0
\end{split}
\end{equation*}
which implies that
\begin{equation*}
\begin{split}
&\eta\sum_{t=1}^T\langle\nabla f_t(\x_t),\x_{t+1}^\ast-\x^\ast\rangle\\
\leq&\|\x^\ast-\x_1\|_2^2-\|\x^\ast_2-\x_1\|_2^2.
\end{split}
\end{equation*}
Since $\|\x^\ast-\x_1\|_2^2\leq D^2$ and $\|\x^\ast_2-\x_1\|_2^2\geq0$, we have
\begin{equation}
\label{thm1_eq2}
\begin{split}
\sum_{t=1}^T\langle\nabla f_t(\x_t),\x_{t+1}^\ast-\x^\ast\rangle\leq\frac{1}{\eta}\|\x^\ast-\x_1\|_2^2\leq\frac{D^2}{\eta}.
\end{split}
\end{equation}
Moreover, by combining the fact that $F_{t}(\x)$ is $2$-strongly convex with (\ref{cor_scvx}), we have
\begin{align*}
&\|\x_t^\ast-\x_{t+1}^\ast\|_2^2\\
\leq& F_t(\x_t^\ast)-F_t(\x_{t+1}^\ast)\\
=&F_{t-1}(\x_t^\ast)-F_{t-1}(\x_{t+1}^\ast)+\eta\nabla f_t(\x_t)^\top(\x_t^\ast-\x_{t+1}^\ast)\\
\leq&\eta\|\nabla f_t(\x_t)\|_2\|\x_t^\ast-\x_{t+1}^\ast\|_2
\end{align*}
which implies that
\begin{equation}
\label{thm1_eq3}
\|\x_t^\ast-\x_{t+1}^\ast\|_2\leq\eta\|\nabla f_t(\x_t)\|_2\leq\eta G.
\end{equation}
By combining (\ref{thm1_eq2}), (\ref{thm1_eq3}) and $\eta=\frac{D}{2G(T+2)^{2/3}}$, we have
\begin{equation}
\label{thm1_eq4}
\begin{split}
B=&\sum_{t=1}^T\langle\nabla f_t(\x_t),\x_t^\ast-\x^\ast\rangle\\
=&\sum_{t=1}^T\langle\nabla f_t(\x_t),\x_{t+1}^\ast-\x^\ast\rangle\\
&+\sum_{t=1}^T\langle\nabla f_t(\x_t),\x_t^\ast-\x_{t+1}^\ast\rangle\\
\leq&\frac{D^2}{\eta}+\sum_{t=1}^T\|\nabla f_t(\x_t)\|_2\|\x_t^\ast-\x_{t+1}^\ast\|_2\\
\leq&\frac{D^2}{\eta}+\eta TG^2\\
\leq&2DG(T+2)^{2/3}+\frac{DG(T+2)^{1/3}}{2}\\
\leq&G\sqrt{C}(T+2)^{2/3}+\frac{G\sqrt{C}(T+2)^{2/3}}{4}
\end{split}
\end{equation}
where the last inequality is due to $D\leq\sqrt{C}/2$ and $(T+2)^{1/3}\leq(T+2)^{2/3}$ for any $T\geq1$.

By combining (\ref{thm1_eq5}) and (\ref{thm1_eq4}), we complete the proof.

\subsection{Proof of Theorem~\ref{thm2}}
Let $\tilde{f}_t(\x)=\langle\nabla f_t(\x_t),\x\rangle+\frac{\lambda}{2}\|\x-\x_t\|_2^2$ for any $t\in[T]$ and  $\x_{t}^\ast=\argmin_{\x\in \K}F_{t-1}(\x)$ for any $t=2,\cdots,T+1$.

Since each function $f_t(\x)$ is $\lambda$-strongly convex, we have
\begin{equation*}
\begin{split}
&\sum_{t=1}^Tf_t(\x_t)-\sum_{t=1}^Tf_t(\x^\ast)\\
\leq& \sum_{t=1}^T\left(\langle\nabla f_t(\x_t),\x_t-\x^\ast\rangle-\frac{\lambda}{2}\|\x_t-\x^\ast\|_2^2\right)\\
=&\sum_{t=1}^T(\f_t(\x_t)-\f_t(\x^\ast))\\
=&\underbrace{\sum_{t=1}^T(\f_t(\x_t)-\f_t(\x_{t+1}^\ast))}_{:=A}+\underbrace{\sum_{t=1}^T(\f_t(\x_{t+1}^\ast)-\f_t(\x^\ast))}_{:=B}.
\end{split}
\end{equation*}
So, we will derive a regret bound by analyzing $A$ and $B$.

To bound $A$, we introduce the following lemma.
\begin{lem}
\label{lem_1_thm2_lem1}
(Lemma 6.7 of \citet{Garber16})
For any $t\in[T]$, the function $\f_t(\x)=\langle\nabla f_t(\x_t),\x\rangle+\frac{\lambda}{2}\|\x-\x_t\|_2^2$ is $(G+\lambda D)$-Lipschitz over $\K$.
\end{lem}
By applying Lemma~\ref{lem_1_thm2_lem1}, for any $t=3,\cdots,T+1$, we have
\begin{equation*}
\begin{split}
&F_{t-1}(\x_{t-1}^\ast)-F_{t-1}(\x_t^\ast)\\
=&F_{t-2}(\x_{t-1}^\ast)-F_{t-2}(\x_t^\ast)+\f_{t-1}(\x_{t-1}^\ast)-\f_{t-1}(\x_{t}^\ast)\\
\leq&(G+\lambda D)\|\x_{t-1}^\ast-\x_{t}^\ast\|_2.
\end{split}
\end{equation*}
Moreover, since each $F_t(\x)$ is $t\lambda$-strongly convex, for any $t=3,\cdots,T+1$, we have
\begin{equation*}
\begin{split}
\|\x_{t-1}^\ast-\x_{t}^\ast\|_2^2\leq&\frac{2(F_{t-1}(\x_{t-1}^\ast)-F_{t-1}(\x_{t}^\ast))}{(t-1)\lambda}\\
\leq&\frac{2(G+\lambda D)\|\x_{t-1}^\ast-\x_{t}^\ast\|_2}{(t-1)\lambda}.
\end{split}
\end{equation*}
Therefore, for any $t=3,\cdots,T+1$, we have
\begin{equation}
\label{pre_eq3}
\begin{split}
\|\x_{t-1}^\ast-\x_{t}^\ast\|_2\leq&\frac{2(G+\lambda D)}{(t-1)\lambda}.
\end{split}
\end{equation}
By applying Lemmas~\ref{thm2_lem1} and~\ref{lem_1_thm2_lem1}, we have
\begin{equation*}
\begin{split}
&\sum_{t=2}^T(\f_t(\x_t)-\f_t(\x_{t+1}^\ast))\\
\leq&\sum_{t=2}^T(G+\lambda D)\|\x_t-\x_{t+1}^\ast\|_2\\
\leq&(G+\lambda D)\sum_{t=2}^T\|\x_t-\x_{t}^\ast\|_2\\
&+(G+\lambda D)\sum_{t=2}^T\|\x_{t}^\ast-\x_{t+1}^\ast\|_2\\
\leq&(G+\lambda D)\sum_{t=2}^T\sqrt{\frac{2(F_{t-1}(\x_{t})-F_{t-1}(\x_{t}^\ast))}{(t-1)\lambda}}\\
&+(G+\lambda D)\sum_{t=2}^T\frac{2(G+\lambda D)}{t\lambda}\\
\leq&(G+\lambda D)\sum_{t=2}^T\sqrt{\frac{2C}{(t-1)\lambda}}+(G+\lambda D)\sum_{t=2}^T\frac{2(G+\lambda D)}{t\lambda}\\
\leq&2(G+\lambda D)\sqrt{\frac{2TC}{\lambda}}+2(G+\lambda D)^2\frac{\ln T}{\lambda}\\
\leq&C\sqrt{2T}+\frac{C\ln T}{2}
\end{split}
\end{equation*}
where the third inequality is due to $\|\x_{t}-\x_{t}^\ast\|_2\leq\sqrt{\frac{2(F_{t-1}(\x_{t})-F_{t-1}(\x_{t}^\ast))}{(t-1)\lambda}}$ for $t\geq2$ and (\ref{pre_eq3}), and the last inequality is due to $\frac{2(G+\lambda D)}{\sqrt{\lambda}}\leq \sqrt{C}$.

Due to $\|\nabla f_1(\x_1)\|_2\leq G$ and $\|\x_1-\x_{2}^\ast\|_2\leq D$, we have
\begin{equation}
\label{thm2_eq3}
\begin{split}
A=&\f_1(\x_1)-\f_1(\x_{2}^\ast)+\sum_{t=2}^T(\f_t(\x_t)-\f_t(\x_{t+1}^\ast))\\
=&\langle\nabla f_1(\x_1),\x_1-\x_2^\ast\rangle-\frac{\lambda}{2}\|\x_2^\ast-\x_1\|_2^2\\
&+\sum_{t=2}^T(\f_t(\x_t)-\f_t(\x_{t+1}^\ast))\\
\leq&\|\nabla f_1(\x_1)\|_2\|\x_1-\x_{2}^\ast\|_2+C\sqrt{2T}+\frac{C\ln T}{2}\\
\leq&GD+C\sqrt{2T}+\frac{C\ln T}{2}.
\end{split}
\end{equation}
Moreover, we note that $F_{t}(\x)=\sum_{\tau=1}^t\f_\tau(\x)$ and $\x_{t+1}^\ast=\argmin_{\x\in \K}F_{t}(\x)$ for any $t\in[T]$. By applying Lemma~\ref{lem_garber16} to $\{\f_t(\x)\}_{t=1}^T$, we have
\begin{equation}
\label{eq2}
\begin{split}
B=\sum_{t=1}^T\f_t(\x_{t+1}^\ast)-\sum_{t=1}^T\f_t(\x^\ast)\leq 0.
\end{split}
\end{equation}
By combining (\ref{thm2_eq3}) and (\ref{eq2}), we complete the proof.

\subsection{Proof of Lemma~\ref{thm1_lem1}}
For brevity, we define $h_{t}=F_{t-1}(\x_{t})-F_{t-1}(\x_t^\ast)$ for $t=1,\cdots,T$ and $h_t(\x_{t-1})=F_{t-1}(\x_{t-1})-F_{t-1}(\x_t^\ast)$ for $t=2,\cdots,T$.

For $t=1$, since $\x_1=\argmin_{\x\in\K}\|\x-\x_1\|_2^2$, we have
\begin{equation}
\label{thm1_lem1_eq1}
h_1=F_{0}(\x_1)-F_{0}(\x_1^\ast)=0\leq\frac{C}{(1+2)^{2/3}}=\epsilon_1.
\end{equation}
For any $T+1\geq t\geq2$, assuming $h_{t-1}\leq\epsilon_{t-1}$ , we first note that
\begin{equation}
\label{lem1_s1_jul}
\begin{split}
&h_t(\x_{t-1})\\
=&F_{t-1}(\x_{t-1})-F_{t-1}(\x_t^\ast)\\
=&F_{t-2}(\x_{t-1})-F_{t-2}(\x_t^\ast)\\
&+\langle\eta\nabla f_{t-1}(\x_{t-1}),\x_{t-1}-\x_t^\ast\rangle\\
\leq& F_{t-2}(\x_{t-1})-F_{t-2}(\x_{t-1}^\ast)\\
&+\langle\eta\nabla f_{t-1}(\x_{t-1}),\x_{t-1}-\x_t^\ast\rangle\\
\leq& \epsilon_{t-1}+\eta\|\nabla f_{t-1}(\x_{t-1})\|_2\|\x_{t-1}-\x_t^\ast\|_2\\
\leq& \epsilon_{t-1}+\eta\|\nabla f_{t-1}(\x_{t-1})\|_2\|\x_{t-1}-\x_{t-1}^\ast\|_2\\
&+\eta\|\nabla f_{t-1}(\x_{t-1})\|_2\|\x_{t-1}^\ast-\x_t^\ast\|_2\\
\leq&\epsilon_{t-1}+\eta G\sqrt{\epsilon_{t-1}}+\eta^2G^2
\end{split}
\end{equation}
where the first inequality is due to $\x_{t-1}^\ast=\argmin_{\x\in\K}F_{t-2}(\x)$ and the last inequality is due to $\|\x_{t-1}-\x_{t-1}^\ast\|_2\leq\sqrt{F_{t-2}(\x_{t-1})-F_{t-2}(\x_{t-1}^\ast)}$ and (\ref{thm1_eq3}).

Then, by substituting $\eta=\frac{D}{2G(T+2)^{2/3}}$ into (\ref{lem1_s1_jul}), we have
\begin{equation}
\label{thm1_lem1_eq2}
\begin{split}
&h_t(\x_{t-1})\\
\leq&\epsilon_{t-1}+\frac{D\sqrt{C}}{2(T+2)^{2/3}(t+1)^{1/3}}+\frac{D^2}{4(T+2)^{4/3}}\\
\leq&\epsilon_{t-1}+\frac{\epsilon_{t-1}}{4(t+1)^{1/3}}+\frac{\epsilon_{t-1}}{16(t+1)^{2/3}}\\
\leq&\left(1+\frac{1}{2(t+1)^{1/3}}\right)\epsilon_{t-1}
\end{split}
\end{equation}
where the second inequality is due to $T\geq t-1$ and $D\leq\frac{\sqrt{C}}{2}$.

Then, to bound $h_t=F_{t-1}(\x_{t})-F_{t-1}(\x_t^\ast)$ by $\epsilon_t$, we further introduce the following lemma.
\begin{lem}
\label{lem_SC_sets}(Derived from Lemma 1 of \citet{Gaber_ICML_15})
Let $f(\x):\K\to\R$ be a convex and $\beta_f$-smooth function, where $\K$ is $\alpha_K$-strongly convex with respect to the $\ell_2$ norm. Moreover, let $\x_{\ii}\in\K$ and $\x_{\oo}=\x_{\ii}+\sigma^\prime(\v-\x_{\ii})$, where $\v\in\argmin_{\mathbf{x}\in\mathcal{K}} \langle\nabla f(\x_{\ii}),\mathbf{x}\rangle$ and $\sigma^\prime=\argmin_{\sigma\in[0,1]}\langle\sigma(\v-\x_{\ii}),\nabla f(\x_{\ii})\rangle+\frac{\sigma^2\beta_{f}}{2}\|\v-\x_{\ii}\|_2^2$. For any $\x^\ast\in\argmin_{\x\in \K}f(\x)$, we have
\begin{align*}
&f(\x_{\oo})-f(\x^\ast)\\
\leq&(f(\x_{\ii})-f(\x^\ast))\max\left(\frac{1}{2},1-\frac{\alpha_K\|\nabla f(\x_{\ii})\|_2}{8\beta_f}\right).
\end{align*}
\end{lem}
We note that $F_{t-1}(\x)$ is $2$-smooth for any $t\in[T+1]$. By applying Lemma~\ref{lem_SC_sets} with $f(\x)=F_{t-1}(\x)$ and $\x_{\ii}=\x_{t-1}$, for any $t\in[T+1]$, we have $\x_{\oo}=\x_{t}$ and
\begin{equation}
\label{SC_pro1}
\begin{split}
h_t\leq&h_t(\x_{t-1})\max\left(\frac{1}{2},1-\frac{\alpha_K\|\nabla F_{t-1}(\x_{t-1})\|_2}{16}\right).
\end{split}
\end{equation}
Because of (\ref{thm1_lem1_eq2}), (\ref{SC_pro1}) and $1+\frac{1}{2(t+1)^{1/3}}\leq\frac{3}{2}$, if $\frac{1}{2}\leq\frac{\alpha_K\|\nabla F_{t-1}(\x_{t-1})\|_2}{16}$, it is easy to verify that
\begin{equation}
\label{thm1_lem1_eq3}
\begin{split}
h_{t}&\leq\frac{3}{4}\epsilon_{t-1}=\frac{3}{4}\frac{C}{(t+1)^{2/3}}\\
&=\frac{C}{(t+2)^{2/3}}\frac{3(t+2)^{2/3}}{4(t+1)^{2/3}}\\
&\leq\frac{C}{(t+2)^{2/3}}=\epsilon_t
\end{split}
\end{equation}
where the last inequality is due to $\frac{3(t+2)^{2/3}}{4(t+1)^{2/3}}\leq1$ for any $t\geq2$.

Then, if $\frac{1}{2}>\frac{\alpha_K\|\nabla F_{t-1}(\x_{t-1})\|_2}{16}$, there exist two cases. First, if $h_t(\x_{t-1})\leq\frac{3C}{4(t+1)^{2/3}}$, it is easy to verify that
\begin{equation}
\label{thm1_lem1_eq4}
h_{t}\leq h_t(\x_{t-1})\leq\frac{3C}{4(t+1)^{2/3}}\leq\epsilon_t
\end{equation}
where the lase inequality has been proved in (\ref{thm1_lem1_eq3}).

Second, if $h_t(\x_{t-1})>\frac{3C}{4(t+1)^{2/3}}$, we have
\begin{equation*}
\begin{split}
&h_t\\
\leq& h_t(\x_{t-1})\left(1-\frac{\alpha_K\|\nabla F_{t-1}(\x_{t-1})\|_2}{16}\right)\\
\leq&\epsilon_{t-1}\left(1+\frac{1}{2(t+1)^{1/3}}\right)\left(1-\frac{\alpha_K\|\nabla F_{t-1}(\x_{t-1})\|_2}{16}\right)\\
\leq&\epsilon_{t-1}\left(1+\frac{1}{2(t+1)^{1/3}}\right)\left(1-\frac{\alpha_K\sqrt{h_t(\x_{t-1})}}{16}\right)\\
\leq&\epsilon_t\frac{(t+2)^{2/3}}{(t+1)^{2/3}}\left(1+\frac{1}{2(t+1)^{1/3}}\right)\left(1-\frac{\alpha_K\sqrt{3C}}{32(t+1)^{1/3}}\right)
\end{split}
\end{equation*}
where the second inequality is due to (\ref{thm1_lem1_eq2}) and the third inequality is due to (\ref{dual_cor_scvx}).

Since $\frac{(t+2)^{2/3}}{(t+1)^{2/3}}\leq 1+\frac{1}{(t+1)^{2/3}}$ for any $t\geq 0$, it is easy to verify that
\begin{align*}
\frac{(t+2)^{2/3}}{(t+1)^{2/3}}\left(1+\frac{1}{2(t+1)^{1/3}}\right)\leq1+\frac{2}{(t+1)^{1/3}}
\end{align*}
which further implies that
\begin{equation}
\label{thm1_lem1_eq5}
\begin{split}
h_t\leq&\epsilon_t\left(1+\frac{2}{(t+1)^{1/3}}\right)\left(1-\frac{\alpha_K\sqrt{3C}}{32(t+1)^{1/3}}\right)\\
\leq&\epsilon_t\left(1+\frac{2}{(t+1)^{1/3}}\right)\left(1-\frac{2}{(t+1)^{1/3}}\right)\\
\leq&\epsilon_t.
\end{split}
\end{equation}
where the second inequality is due to $\alpha_K\sqrt{3C}\geq64$.

By combining (\ref{thm1_lem1_eq1}), (\ref{thm1_lem1_eq3}), (\ref{thm1_lem1_eq4}) and (\ref{thm1_lem1_eq5}), we complete the proof.

\section{Conclusion and Future Work}
In this paper, we first prove that the classical OFW algorithm with line search attains an $O(T^{2/3})$ regret bound for OCO over strongly convex sets, which is better than the $O(T^{3/4})$ regret bound for the general OCO. Furthermore, for strongly convex losses, we introduce a strongly convex variant of OFW, and prove that it achieves a regret bound of $O(T^{2/3})$ over general convex sets and a better regret bound of $O(\sqrt{T})$ over strongly convex sets.

An open question is whether the regret of OFW and its strongly convex variant over strongly convex sets can
be further improved if the losses are smooth. We note that \citet{Hazan20} have proposed a projection-free algorithm for OCO over general convex sets, and established an improved regret bound of $O(T^{2/3})$ by taking advantage of the smoothness.

\section{Acknowledgments}
This work was partially supported by the NSFC (61921006), NSFC-NRF Joint Research Project (61861146001), and the Collaborative Innovation Center of Novel Software Technology and Industrialization.

\bibliography{ref}

\begin{thebibliography}{37}
\providecommand{\natexlab}[1]{#1}
\providecommand{\url}[1]{\texttt{#1}}
\providecommand{\urlprefix}{URL }
\expandafter\ifx\csname urlstyle\endcsname\relax
  \providecommand{\doi}[1]{doi:\discretionary{}{}{}#1}\else
  \providecommand{\doi}{doi:\discretionary{}{}{}\begingroup
  \urlstyle{rm}\Url}\fi

\bibitem[{Agarwal et~al.(2006)Agarwal, Hazan, Kale, and
  Schapire}]{Portfolio_Agarwal}
Agarwal, A.; Hazan, E.; Kale, S.; and Schapire, R.~E. 2006.
\newblock Algorithms for portfolio management based on the {N}ewton method.
\newblock In \emph{Proceedings of the 23rd International Conference on Machine
  Learning}, 9--16.

\bibitem[{Awerbuch and Kleinberg(2008)}]{Awerbuch2008}
Awerbuch, B.; and Kleinberg, R. 2008.
\newblock Online linear optimization and adaptive routing.
\newblock \emph{Journal of Computer and System Sciences} 74(1): 97--114.

\bibitem[{Awerbuch and Kleinberg(2004)}]{Awerbuch04}
Awerbuch, B.; and Kleinberg, R.~D. 2004.
\newblock Adaptive routing with end-to-end feedback: Distributed learning and
  geometric approaches.
\newblock In \emph{Proceedings of the 36th Annual ACM Symposium on Theory of
  Computing}, 45--53.

\bibitem[{Blum and Kalai(1999)}]{Blum1999}
Blum, A.; and Kalai, A. 1999.
\newblock Universal portfolios with and without transaction costs.
\newblock \emph{Machine Learning} 35(3): 193--205.

\bibitem[{Boyd and Vandenberghe(2004)}]{Boyd04}
Boyd, S.; and Vandenberghe, L. 2004.
\newblock \emph{Convex Optimization}.
\newblock Cambridge University Press.

\bibitem[{Bubeck et~al.(2015)Bubeck, Dekel, Koren, and Peres}]{Bubeck15}
Bubeck, S.; Dekel, O.; Koren, T.; and Peres, Y. 2015.
\newblock Bandit convex optimization: $\sqrt{T}$ regret in one dimension.
\newblock In \emph{Proceedings of the 28th Conference on Learning Theory},
  266--278.

\bibitem[{Chen et~al.(2016)Chen, Yang, Lin, Zhang, and Chang}]{chen-UAI2016}
Chen, J.; Yang, T.; Lin, Q.; Zhang, L.; and Chang, Y. 2016.
\newblock Optimal stochastic strongly convex optimization with a logarithmic
  number of projections.
\newblock In \emph{Proceedings of the 32nd Conference on Uncertainty in
  Artificial Intelligence}, 122--131.

\bibitem[{Chen, Zhang, and Karbasi(2019)}]{chen19}
Chen, L.; Zhang, M.; and Karbasi, A. 2019.
\newblock Projection-free bandit convex optimization.
\newblock In \emph{Proceedings of the 22nd International Conference on
  Artificial Intelligence and Statistics}, 2047--2056.

\bibitem[{Demyanov and Rubinov(1970)}]{Demyanov70}
Demyanov, V.~F.; and Rubinov, A.~M. 1970.
\newblock \emph{Approximate Methods in Optimization Problems}.
\newblock Elsevier Publishing Company.

\bibitem[{Duchi, Agarwal, and Wainwright(2011)}]{DADO2011}
Duchi, J.~C.; Agarwal, A.; and Wainwright, M.~J. 2011.
\newblock Dual averaging for distributed optimization: Convergence analysis and
  network scaling.
\newblock \emph{IEEE Transactions on Automatic Control} 57(3): 592--606.

\bibitem[{Dunn(1979)}]{Dunn79}
Dunn, J.~C. 1979.
\newblock Rates of convergence for conditional gradient algorithms near
  singular and nonsingular extremals.
\newblock \emph{SIAM Journal on Control and Optimization} 17(2): 187--211.

\bibitem[{Flaxman, Kalai, and McMahan(2005)}]{OBO05}
Flaxman, A.~D.; Kalai, A.~T.; and McMahan, H.~B. 2005.
\newblock Online convex optimization in the bandit setting: Gradient descent
  without a gradient.
\newblock In \emph{Proceedings of the 16th Annual ACM-SIAM Symposium on
  Discrete Algorithms}, 385--394.

\bibitem[{Frank and Wolfe(1956)}]{FW-56}
Frank, M.; and Wolfe, P. 1956.
\newblock An algorithm for quadratic programming.
\newblock \emph{Naval Research Logistics Quarterly} 3(1--2): 95--110.

\bibitem[{Garber and Hazan(2015)}]{Gaber_ICML_15}
Garber, D.; and Hazan, E. 2015.
\newblock Faster rates for the frank-wolfe method over strongly-convex sets.
\newblock In \emph{Proceedings of the 32nd International Conference on Machine
  Learning}, 541--549.

\bibitem[{Garber and Hazan(2016)}]{Garber16}
Garber, D.; and Hazan, E. 2016.
\newblock A linearly convergent conditional gradient algorithm with
  applications to online and stochastic optimization.
\newblock \emph{SIAM Journal on Optimization} 26(3): 1493--1528.

\bibitem[{Garber and Kretzu(2020{\natexlab{a}})}]{Garber19}
Garber, D.; and Kretzu, B. 2020{\natexlab{a}}.
\newblock Improved regret bounds for projection-free bandit convex
  optimization.
\newblock In \emph{Proceedings of the 23rd International Conference on
  Artificial Intelligence and Statistics}, 2196--2206.

\bibitem[{Garber and Kretzu(2020{\natexlab{b}})}]{Garber_SOFW}
Garber, D.; and Kretzu, B. 2020{\natexlab{b}}.
\newblock Revisiting projection-free online learning: the strongly convex case.
\newblock \emph{ArXiv e-prints} arXiv: 2010.07572.

\bibitem[{Hazan(2008)}]{FW_Hazan08}
Hazan, E. 2008.
\newblock Sparse approximate solutions to semidefinite programs.
\newblock In \emph{Latin American Symposium on Theoretical Informatics},
  306--316.

\bibitem[{Hazan(2016)}]{Hazan2016}
Hazan, E. 2016.
\newblock Introduction to online convex optimization.
\newblock \emph{Foundations and Trends in Optimization} 2(3--4): 157--325.

\bibitem[{Hazan, Agarwal, and Kale(2007)}]{Hazan_2007}
Hazan, E.; Agarwal, A.; and Kale, S. 2007.
\newblock Logarithmic regret algorithms for online convex optimization.
\newblock \emph{Machine Learning} 69(2): 169--192.

\bibitem[{Hazan and Kale(2012)}]{Hazan2012}
Hazan, E.; and Kale, S. 2012.
\newblock Projection-free online learning.
\newblock In \emph{Proceedings of the 29th International Conference on Machine
  Learning}, 1843--1850.

\bibitem[{Hazan and Luo(2016)}]{SVFW}
Hazan, E.; and Luo, H. 2016.
\newblock Variance-reduced and projection-free stochastic optimization.
\newblock In \emph{Proceedings of the 33rd International Conference on Machine
  Learning}, 1263--1271.

\bibitem[{Hazan and Minasyan(2020)}]{Hazan20}
Hazan, E.; and Minasyan, E. 2020.
\newblock Faster projection-free online learning.
\newblock In \emph{Proceedings of the 33rd Annual Conference on Learning
  Theory}, 1877--1893.

\bibitem[{Hosseini, Chapman, and Mesbahi(2013)}]{D-ODA}
Hosseini, S.; Chapman, A.; and Mesbahi, M. 2013.
\newblock Online distributed optimization via dual averaging.
\newblock In \emph{52nd IEEE Conference on Decision and Control}, 1484--1489.

\bibitem[{Jaggi(2013)}]{Revist_FW}
Jaggi, M. 2013.
\newblock Revisiting frank-wolfe: Projection-free sparse convex optimization.
\newblock In \emph{Proceedings of the 30th International Conference on Machine
  Learning}, 427--435.

\bibitem[{Levitin and Polyak(1966)}]{Levitin66}
Levitin, E.~S.; and Polyak, B.~T. 1966.
\newblock Constrained minimization methods.
\newblock \emph{USSR Computational mathematics and mathematical physics} 6:
  1--50.

\bibitem[{Levy and Krause(2019)}]{kevy_smooth}
Levy, K.~Y.; and Krause, A. 2019.
\newblock Projection free online learning over smooth sets.
\newblock In \emph{Proceedings of the 22nd International Conference on
  Artificial Intelligence and Statistics}, 1458--1466.

\bibitem[{Luo, Wei, and Zheng(2018)}]{Portfolio_Luo}
Luo, H.; Wei, C.-Y.; and Zheng, K. 2018.
\newblock Efficient online portfolio with logarithmic regret.
\newblock In \emph{Advances in Neural Information Processing Systems 31},
  8235--8245.

\bibitem[{Rector-Brooks, Wang, and Mozafari(2019)}]{Jarrid19}
Rector-Brooks, J.; Wang, J.-K.; and Mozafari, B. 2019.
\newblock Revisiting projection-free optimization for strongly convex
  constraint sets.
\newblock In \emph{Proceedings of the 33rd AAAI Conference on Artificial
  Intelligence}, 1576--1583.

\bibitem[{Shalev-Shwartz(2007)}]{ShaiThesis}
Shalev-Shwartz, S. 2007.
\newblock \emph{Online Learning: Theory, Algorithms, and Applications}.
\newblock Ph.D. thesis, The Hebrew University of Jerusalem.

\bibitem[{Shalev-Shwartz(2011)}]{Online:suvery}
Shalev-Shwartz, S. 2011.
\newblock Online learning and online convex optimization.
\newblock \emph{Foundations and Trends in Machine Learning} 4(2): 107--194.

\bibitem[{Shalev-Shwartz and Singer(2007)}]{Shai07}
Shalev-Shwartz, S.; and Singer, Y. 2007.
\newblock A primal-dual perspective of online learning algorithm.
\newblock \emph{Machine Learning} 69(2--3): 115--142.

\bibitem[{Wan, Tu, and Zhang(2020)}]{Wan-ICML-2020}
Wan, Y.; Tu, W.-W.; and Zhang, L. 2020.
\newblock Projection-free distributed online convex optimization with
  ${O}(\sqrt{T})$ communication complexity.
\newblock In \emph{Proceedings of the 37th International Conference on Machine
  Learning}, 9818--9828.

\bibitem[{Yang, Lin, and Zhang(2017)}]{Yang-ICML17}
Yang, T.; Lin, Q.; and Zhang, L. 2017.
\newblock A richer theory of convex constrained optimization with reduced
  projections and improved rates.
\newblock In \emph{Proceedings of the 34th International Conference on Machine
  Learning}, 3901--3910.

\bibitem[{Zhang et~al.(2013)Zhang, Yang, Jin, and He}]{Zhang-ICML13}
Zhang, L.; Yang, T.; Jin, R.; and He, X. 2013.
\newblock {$O(\log T)$} projections for stochastic optimization of smooth and
  strongly convex functions.
\newblock In \emph{Proceedings of the 30th International Conference on Machine
  Learning}, 1121--1129.

\bibitem[{Zhang et~al.(2017)Zhang, Zhao, Zhu, Hoi, and Zhang}]{wenpeng17}
Zhang, W.; Zhao, P.; Zhu, W.; Hoi, S. C.~H.; and Zhang, T. 2017.
\newblock Projection-free distributed online learning in networks.
\newblock In \emph{Proceedings of the 34th International Conference on Machine
  Learning}, 4054--4062.

\bibitem[{Zinkevich(2003)}]{Zinkevich2003}
Zinkevich, M. 2003.
\newblock Online convex programming and generalized infinitesimal gradient
  ascent.
\newblock In \emph{Proceedings of the 20th International Conference on Machine
  Learning}, 928--936.

\end{thebibliography}

\clearpage
\onecolumn

\section{Proof of Lemma~\ref{thm2_lem1}}
Let $\tilde{f}_t(\x)=\langle\nabla f_t(\x_t),\x\rangle+\frac{\lambda}{2}\|\x-\x_t\|_2^2$ for $t=1,\cdots,T$. For brevity, we define $h_{t}=F_{t-1}(\x_{t})-F_{t-1}(\x_t^\ast)$ for $t=2,\cdots,T$ and $h_t(\x_{t-1})=F_{t-1}(\x_{t-1})-F_{t-1}(\x_t^\ast)$ for $t=3,\cdots,T$. For $t=2$, we have
\begin{equation}
\label{thm2_lem1_eq1}
\begin{split}
h_2=&F_{1}(\x_2)-F_{1}(\x_2^\ast)\\
=&\langle\nabla f_{1}(\x_1),\x_2\rangle+\frac{\lambda}{2}\|\x_2-\x_1\|_2^2-\langle\nabla f_{1}(\x_1),\x_2^\ast\rangle-\frac{\lambda}{2}\|\x_2^\ast-\x_1\|_2^2\\
\leq&\langle\nabla f_{1}(\x_1),\x_2-\x_2^\ast\rangle+\frac{\lambda}{2}\|\x_2-\x_1\|_2^2\\
\leq&\|\nabla f_{1}(\x_1)\|_2\|\x_2-\x_2^\ast\|_2+\frac{\lambda}{2}\|\x_2-\x_1\|_2^2\\
\leq& GD+\frac{\lambda D^2}{2}\\
\leq& C.
\end{split}
\end{equation}
For any $T+1\geq t\geq3$, assuming $h_{t-1}\leq C$, we have
\begin{equation*}
\begin{split}
&h_t(\x_{t-1})\\
=&F_{t-1}(\x_{t-1})-F_{t-1}(\x_t^\ast)\\
=&F_{t-2}(\x_{t-1})-F_{t-2}(\x_t^\ast)+\f_{t-1}(\x_{t-1})-\f_{t-1}(\x_{t}^\ast)\\
\leq& F_{t-2}(\x_{t-1})-F_{t-2}(\x_{t-1}^\ast)+(G+\lambda D)\|\x_{t-1}-\x_t^\ast\|_2\\
\leq& C+(G+\lambda D)\|\x_{t-1}-\x_{t-1}^\ast\|_2+(G+\lambda D)\|\x_{t-1}^\ast-\x_t^\ast\|_2\\
\leq&C+(G+\lambda D)\sqrt{\frac{2C}{(t-2)\lambda}}+\frac{2(G+\lambda D)^2}{(t-1)\lambda}
\end{split}
\end{equation*}
where the first inequality is due to Lemma~\ref{lem_1_thm2_lem1} and the last inequality is due to (\ref{pre_eq3}) and \begin{equation}
\label{thm2_thm3_eqx}
\|\x_{t-1}-\x_{t-1}^\ast\|_2\leq\sqrt{\frac{2(F_{t-2}(\x_{t-1})-F_{t-2}(\x_{t-1}^\ast))}{(t-2)\lambda}}.
\end{equation}
Let $c_t=1+\frac{1}{\sqrt{t-1}}+\frac{1}{2(t-1)}$. Because of $\frac{2(G+\lambda D)}{\sqrt{\lambda}}\leq \sqrt{C}$, we further have
\begin{equation*}
\begin{split}
h_t(\x_{t-1})
\leq&C+\frac{C}{2}\sqrt{\frac{2}{(t-2)}}+\frac{C}{2(t-1)}\\
\leq&C+\frac{C}{2}\sqrt{\frac{4}{t-1}}+\frac{C}{2(t-1)}\\
\leq&C\left(1+\frac{1}{\sqrt{t-1}}+\frac{1}{2(t-1)}\right)\\
=&c_tC
\end{split}
\end{equation*}
where the second inequality is due to $\frac{1}{t-2}\leq\frac{2}{t-1}$ for any $t\geq 3$.

We note that $F_{t-1}(\x)$ is $(t-1)\lambda$-smooth for any $t=2,\cdots,T+1$. By applying Lemma~\ref{lem_SC_sets} with $f(\x)=F_{t-1}(\x)$ and $\x_{\ii}=\x_{t-1}$, we have $\x_{\oo}=\x_{t}$ and
\begin{equation}
\label{SC_sets_pro2}
h_t\leq h_t(\x_{t-1})\max\left(\frac{1}{2},1-\frac{\alpha_K\|\nabla F_{t-1}(\x_{t-1})\|_2}{8(t-1)\lambda}\right).
\end{equation}
Because of (\ref{SC_sets_pro2}), if $\frac{1}{2}\leq\frac{\alpha_K\|\nabla F_{t-1}(\x_{t-1})\|_2}{8(t-1)\lambda}$, it is easy to verify that
\begin{equation}
\label{thm2_lem1_eq3}
h_{t}\leq\frac{h_t(\x_{t-1})}{2}\leq\frac{c_t}{2}C\leq C
\end{equation}
where the last inequality is due to $c_t\leq2$ for any $t\geq3$.

Otherwise, if $\frac{1}{2}>\frac{\alpha_K\|\nabla F_{t-1}(\x_{t-1})\|_2}{8(t-1)\lambda}$, there exist two cases. First, if $h_t(\x_{t-1})\leq C$, it is easy to verify that
\begin{equation}
\label{thm2_lem1_eq4}
h_{t}\leq h_t(\x_{t-1})\leq C.
\end{equation}
Second, if $c_tC\geq h_t(\x_{t-1})>C$, we have
\begin{equation*}
\begin{split}
h_t\leq& h_t(\x_{t-1})\left(1-\frac{\alpha_K\|\nabla F_{t-1}(\x_{t-1})\|_2}{8(t-1)\lambda}\right)\\
\leq&c_tC\left(1-\frac{\alpha_K\|\nabla F_{t-1}(\x_{t-1})\|_2}{8(t-1)\lambda}\right)\\
\leq&c_tC\left(1-\frac{\alpha_K\sqrt{(t-1)\lambda}\sqrt{h_t(\x_{t-1})}}{8\sqrt{2}(t-1)\lambda}\right)\\
\leq&c_tC\left(1-\frac{\alpha_K\sqrt{C}}{8\sqrt{2}\sqrt{(t-1)\lambda}}\right).
\end{split}
\end{equation*}
where the third inequality is due to (\ref{dual_cor_scvx}).

Because of $c_t=1+\frac{1}{\sqrt{t-1}}+\frac{1}{2(t-1)}\leq1+\frac{3}{2\sqrt{t-1}}$ for any $t\geq 3$ and $\alpha_K\sqrt{C}\geq12\sqrt{2\lambda}$, we further have
\begin{equation}
\label{thm2_lem1_eq5}
h_{t}\leq C\left(1+\frac{3}{2\sqrt{t-1}}\right)\left(1-\frac{3}{2\sqrt{t-1}}\right)\leq C.
\end{equation}
By combining (\ref{thm2_lem1_eq1}), (\ref{thm2_lem1_eq3}), (\ref{thm2_lem1_eq4}) and (\ref{thm2_lem1_eq5}), we complete the proof.

\section{Proof of Lemma~\ref{lem_1_thm2_lem1}}
This lemma has been proved by \citet{Garber16}, we include the proof for completeness.
For any $\x,\y\in\K$ and $t\in[T]$, we have
\begin{align*}
\f_t(\x)-\f_t(\y)&\leq \langle\nabla \f_t(\x),\x-\y\rangle=\langle\nabla f_t(\x_t)+\lambda(\x-\x_t),\x-\y\rangle\\
&\leq\|\nabla f_t(\x_t)+\lambda(\x-\x_t)\|_2\|\x-\y\|_2\leq (G+\lambda D)\|\x-\y\|_2.
\end{align*}
Following the above inequality, we also have
\[\f_t(\y)-\f_t(\x)\leq(G+\lambda D)\|\x-\y\|_2\]
which completes the proof.
\section{Proof of Theorem~\ref{thm3}}
This proof is similar to that of Theorem~\ref{thm2}. Let $\tilde{f}_t(\x)=\langle\nabla f_t(\x_t),\x\rangle+\frac{\lambda}{2}\|\x-\x_t\|_2^2$ for any $t\in[T]$. Let $\x_{t}^\ast=\argmin_{\x\in \K}F_{t-1}(\x)$ for any $t=2,\cdots,T+1$. From the proof of Theorem~\ref{thm2}, we have
\begin{align*}
\sum_{t=1}^Tf_t(\x_t)-\sum_{t=1}^Tf_t(\x^\ast)&\leq A + B
\end{align*}
where $A=\sum_{t=1}^T(\f_t(\x_t)-\f_t(\x_{t+1}^\ast))$ and $B=\sum_{t=1}^T(\f_t(\x_{t+1}^\ast)-\f_t(\x^\ast))$.

Moreover, as in (\ref{eq2}), we have proved that
\[B=\sum_{t=1}^T(\f_t(\x_{t+1}^\ast)-\f_t(\x^\ast))\leq0.\]
Therefore, we only need to bound $A$. For SC-OFW over general convex sets, we introduce a new lemma.
\begin{lem}
\label{thm3_lem1}
Let $\x_{t}^\ast=\argmin_{\x\in \K}F_{t-1}(\x)$ for any $t=2,\cdots,T+1$, where $F_t(\x)$ is defined in (\ref{def_F2}). Then, for any $t=2,\cdots,T+1$, Algorithm~\ref{SC_OFW} has
\[F_{t-1}(\x_{t})-F_{t-1}(\x_t^\ast)\leq \epsilon_t=C(t-1)^{1/3}\]
where $C=\frac{16(G+\lambda D)^2}{\lambda}.$
\end{lem}
By applying Lemmas~\ref{lem_1_thm2_lem1} and~\ref{thm3_lem1}, we have
\begin{equation}
\label{thm3_eq3_pre1}
\begin{split}
\sum_{t=2}^T(\f_t(\x_t)-\f_t(\x_{t+1}^\ast))\leq&\sum_{t=2}^T(G+\lambda D)\|\x_t-\x_{t+1}^\ast\|_2\\
\leq&(G+\lambda D)\sum_{t=2}^T\|\x_t-\x_{t}^\ast\|_2+(G+\lambda D)\sum_{t=2}^T\|\x_{t}^\ast-\x_{t+1}^\ast\|_2\\
\leq&(G+\lambda D)\sum_{t=2}^T\sqrt{\frac{2(F_{t-1}(\x_{t})-F_{t-1}(\x_{t}^\ast))}{(t-1)\lambda}}+(G+\lambda D)\sum_{t=2}^T\frac{2(G+\lambda D)}{t\lambda}\\
\leq&(G+\lambda D)\sum_{t=2}^T\sqrt{\frac{2C(t-1)^{1/3}}{(t-1)\lambda}}+(G+\lambda D)\sum_{t=2}^T\frac{2(G+\lambda D)}{t\lambda}\\
\leq&(G+\lambda D)\sqrt{\frac{2C}{\lambda}}\frac{3}{2}T^{2/3}+2(G+\lambda D)^2\frac{\ln T}{\lambda}\\
=&\frac{3\sqrt{2}CT^{2/3}}{8}+\frac{C\ln T}{8}
\end{split}
\end{equation}
where the third inequality is due to $\|\x_{t}-\x_{t}^\ast\|_2\leq\sqrt{\frac{2(F_{t-1}(\x_{t})-F_{t-1}(\x_{t}^\ast))}{(t-1)\lambda}}$ for $t\geq2$ and (\ref{pre_eq3}), and the last equality is due to $C=\frac{16(G+\lambda D)^2}{\lambda}.$

Because of $\|\nabla f_1(\x_1)\|_2\leq G$ and $\|\x_1-\x_{2}^\ast\|_2\leq D$, we further have
\begin{equation*}
\begin{split}
A=&\f_1(\x_1)-\f_1(\x_{2}^\ast)+\sum_{t=2}^T(\f_t(\x_t)-\f_t(\x_{t+1}^\ast))\\
=&\langle\nabla f_1(\x_1),\x_1-\x_2^\ast\rangle-\frac{\lambda}{2}\|\x_2^\ast-\x_1\|_2^2+\sum_{t=2}^T(\f_t(\x_t)-\f_t(\x_{t+1}^\ast))\\
\leq&\|\nabla f_1(\x_1)\|_2\|\x_1-\x_{2}^\ast\|_2+\frac{3\sqrt{2}CT^{2/3}}{8}+\frac{C\ln T}{8}\\
\leq&GD+\frac{3\sqrt{2}CT^{2/3}}{8}+\frac{C\ln T}{8}
\end{split}
\end{equation*}
which completes this proof.
\section{Proof of Lemma~\ref{thm3_lem1}}
For brevity, we define $h_{t}=F_{t-1}(\x_{t})-F_{t-1}(\x_t^\ast)$ and $\tilde{f}_t(\x)=\langle\nabla f_t(\x_t),\x\rangle+\frac{\lambda}{2}\|\x-\x_t\|_2^2$ for any $t\in[T]$. For $t=2$, following (\ref{thm2_lem1_eq1}), we have
\begin{equation}
\label{thm3_lem1_eq1}
\begin{split}
h_2=F_{1}(\x_2)-F_{1}(\x_2^\ast)\leq GD+\frac{\lambda D^2}{2}\leq C= \epsilon_2.
\end{split}
\end{equation}
For any $T+1\geq t\geq3$, assuming $h_{t-1}\leq \epsilon_{t-1}$ , we have
\begin{equation*}
\begin{split}
&F_{t-1}(\x_{t-1})-F_{t-1}(\x_t^\ast)\\
=&F_{t-2}(\x_{t-1})-F_{t-2}(\x_t^\ast)+\f_{t-1}(\x_{t-1})-\f_{t-1}(\x_{t}^\ast)\\
\leq& F_{t-2}(\x_{t-1})-F_{t-2}(\x_{t-1}^\ast)+(G+\lambda D)\|\x_{t-1}-\x_t^\ast\|_2\\
\leq& \epsilon_{t-1}+(G+\lambda D)\|\x_{t-1}-\x_{t-1}^\ast\|_2+(G+\lambda D)\|\x_{t-1}^\ast-\x_t^\ast\|_2\\
\leq&\epsilon_{t-1}+(G+\lambda D)\sqrt{\frac{2\epsilon_{t-1}}{(t-2)\lambda}}+\frac{2(G+\lambda D)^2}{(t-1)\lambda}
\end{split}
\end{equation*}
where the first inequality is due to Lemma~\ref{lem_1_thm2_lem1} and the last inequality is due to (\ref{pre_eq3}) and (\ref{thm2_thm3_eqx}).

Because of $C=\frac{16(G+\lambda D)^2}{\lambda}$ and $\frac{1}{t-2}\leq\frac{2}{t-1}$ for any $t\geq 3$, we further have
\begin{equation}
\label{thm3_lem1_eq2}
\begin{split}
F_{t-1}(\x_{t-1})-F_{t-1}(\x_t^\ast)\leq&\epsilon_{t-1}+2(G+\lambda D)\sqrt{\frac{\epsilon_{t-1}}{(t-1)\lambda}}+\frac{2(G+\lambda D)^2}{(t-1)\lambda}\\
\leq&C(t-2)^{1/3}+\frac{\sqrt{C\lambda}}{2}\sqrt{\frac{C(t-2)^{1/3}}{(t-1)\lambda}}+\frac{C}{8(t-1)}\\
\leq&C(t-1)^{1/3}+\frac{C}{2(t-1)^{1/3}}+\frac{C}{8(t-1)}.
\end{split}
\end{equation}
Then, for any $\sigma\in[0,1]$, we have
\begin{equation}
\label{thm3_lem1_eq3}
\begin{split}
h_t&=F_{t-1}(\x_{t})-F_{t-1}(\x_t^\ast)=F_{t-1}(\x_{t-1}+\sigma_{t-1}(\mathbf{v}_{t-1}-\mathbf{x}_{t-1}))-F_{t-1}(\x_t^\ast)\\
&\leq F_{t-1}(\x_{t-1})-F_{t-1}(\x_t^\ast)+\sigma_{t-1}\langle\nabla F_{t-1}(\x_{t-1}),\mathbf{v}_{t-1}-\mathbf{x}_{t-1}\rangle+\frac{(t-1)\lambda\sigma_{t-1}^2}{2}\|\mathbf{v}_{t-1}-\mathbf{x}_{t-1}\|_2^2\\
&\leq F_{t-1}(\x_{t-1})-F_{t-1}(\x_t^\ast)+\sigma\langle\nabla F_{t-1}(\x_{t-1}),\mathbf{v}_{t-1}-\mathbf{x}_{t-1}\rangle+\frac{(t-1)\lambda\sigma^2}{2}\|\mathbf{v}_{t-1}-\mathbf{x}_{t-1}\|_2^2\\
&\leq F_{t-1}(\x_{t-1})-F_{t-1}(\x_t^\ast)+\sigma\langle\nabla F_{t-1}(\x_{t-1}),\mathbf{x}_{t}^\ast-\mathbf{x}_{t-1}\rangle+\frac{(t-1)\lambda\sigma^2}{2}\|\mathbf{v}_{t-1}-\mathbf{x}_{t-1}\|_2^2\\
&\leq F_{t-1}(\x_{t-1})-F_{t-1}(\x_t^\ast)+\sigma(F_{t-1}(\x_t^\ast)-F_{t-1}(\x_{t-1}))+\frac{(t-1)\lambda\sigma^2}{2}\|\mathbf{v}_{t-1}-\mathbf{x}_{t-1}\|_2^2\\
&\leq(1-\sigma)(F_{t-1}(\x_{t-1})-F_{t-1}(\x_t^\ast))+\frac{(t-1)\lambda\sigma^2D^2}{2}
\end{split}
\end{equation}
where the first inequality is due to the smoothness of $F_{t-1}(\x)$, the second inequality is due to the line search and the third inequality is due to $\mathbf{v}_{t-1}\in\argmin_{\mathbf{x}\in\mathcal{K}} \langle\nabla F_{t-1}(\x_{t-1}),\mathbf{x}\rangle$.

Because of $0\leq(t-1)^{-2/3}\leq1$ for any $t\geq3$, we substitute $\sigma=(t-1)^{-2/3}$ into (\ref{thm3_lem1_eq3}) and combine it with (\ref{thm3_lem1_eq2}), which implies that
\begin{equation*}
\begin{split}
h_t\leq\left(1-\frac{1}{(t-1)^{2/3}}\right)\left(C(t-1)^{1/3}+\frac{C}{2(t-1)^{1/3}}+\frac{C}{8(t-1)}\right)+\frac{(t-1)\lambda D^2}{2(t-1)^{4/3}}.
\end{split}
\end{equation*}
Then, due to $16\lambda D^2\leq C$, we have
\begin{equation*}
\begin{split}
h_t\leq\left(1-\frac{1}{(t-1)^{2/3}}\right)\left(C(t-1)^{1/3}+\frac{C}{2(t-1)^{1/3}}+\frac{C}{8(t-1)}\right)+\frac{C}{32(t-1)^{1/3}}.
\end{split}
\end{equation*}
Moreover, due to $\frac{1}{4}\leq1-\frac{1}{(t-1)^{2/3}}$ for any $t\geq3$, we have
\begin{equation}
\label{thm3_lem1_eq4}
\begin{split}
h_t&\leq\left(1-\frac{1}{(t-1)^{2/3}}\right)\left(C(t-1)^{1/3}+\frac{C}{2(t-1)^{1/3}}+\frac{C}{8(t-1)^{1/3}}+\frac{C}{8(t-1)}\right)\\
&=\left(1-\frac{1}{(t-1)^{2/3}}\right)\left(1+\frac{5}{8(t-1)^{2/3}}+\frac{C}{8(t-1)^{4/3}}\right)C(t-1)^{1/3}\\
&\leq\left(1-\frac{1}{(t-1)^{2/3}}\right)\left(1+\frac{1}{(t-1)^{2/3}}\right)C(t-1)^{1/3}\\
&\leq \epsilon_t
\end{split}
\end{equation}
By combining (\ref{thm3_lem1_eq1}) and (\ref{thm3_lem1_eq4}), we complete the proof.
\end{document}